\documentclass{ifacconf}

\usepackage{graphicx}      % include this line if your document contains figures
\usepackage{natbib}        % required for bibliography

\usepackage{amssymb}
\usepackage{amsmath}

\newcommand{\qvec}[1]{\mathbf{#1}}
\newcommand{\qmat}[1]{\mathbf{#1}}
\newcommand{\idx}[1]{_{\mathrm{#1}}}

\newcommand{\qRealNumbers}{\mathbb{R}}
\newcommand{\qPositiveNaturalNumbers}{\mathbb{N}_{\geq 0}}
\newcommand{\qNaturalNumbers}{\mathbb{N}}

\newcommand{\qyest}{\hat{\qy}}

\newcommand{\qgrad}{\boldsymbol{\nabla}}
\newcommand{\qdeltau}{\boldsymbol{\Delta}\qu}

\newcommand{\qyvec}{\bar{\qvec{y}}}
\newcommand{\quvec}{\bar{\qvec{u}}}
\newcommand{\qxvec}{\bar{\qvec{x}}}
\newcommand{\qy}{\qvec{y}}
\newcommand{\qu}{\qvec{u}}

\newcommand{\qf}{\qvec{f}}
\newcommand{\qr}{\qvec{r}}
\newcommand{\qe}{\qvec{e}}
\newcommand{\qm}{\qvec{m}}
\newcommand{\qv}{\qvec{v}}

\newcommand{\qC}{\qmat{C}}

\newcommand{\qI}{\qmat{I}}
\newcommand{\qL}{\qmat{L}}
\newcommand{\qP}{\qmat{P}}
\newcommand{\qS}{\qmat{S}}
\newcommand{\qW}{\qmat{W}}

\newcommand{\qNorm}[1]{\left\| {#1} \right\|}

\usepackage{color}

\begin{document}
\begin{frontmatter}

\title{Autonomous Iterative Motion Learning (AI-MOLE) of a SCARA Robot for Automated Myocardial Injection} 
% Title, preferably not more than 10 words.

\thanks[footnoteinfo]{This work was partially funded by the Bundesministerium für Bildung und Forschung within the project TACTiC and grand ID 01EK2108E.}

\author[imes]{Michael Meindl} 
\author[imes]{Raphael Mönkemöller} 
\author[imes]{Thomas Seel}

\address[imes]{Institute of Mechatronic Systems, Leibniz University Hannover, Garbsen, 30823 Germany,  (e-mail: \{michael.meindl,
raphael.moenkemoeller,
thomas.seel\}@imes.uni-hannover).}

\begin{abstract}                % Abstract of 50--100 words
Stem cell therapy is a promising approach to treat heart insufficiency and benefits from automated myocardial injection which requires highly precise motion of a robotic manipulator that is equipped with a syringe.
This work investigates whether sufficiently precise motion can be achieved by combining a SCARA robot and learning control methods.
For this purpose, the method Autonomous Iterative Motion Learning (AI-MOLE) is extended to be applicable to multi-input/multi-output systems.
The proposed learning method solves reference tracking tasks in systems with unknown, nonlinear, multi-input/multi-output dynamics by iteratively updating an input trajectory in a plug-and-play fashion and without requiring manual parameter tuning.
The proposed learning method is validated in a preliminary simulation study of a simplified SCARA robot that has to perform three desired motions.
The results demonstrate that the proposed learning method achieves highly precise reference tracking without requiring any a priori model information or manual parameter tuning in as little as 15 trials per motion.
The results further indicate that the combination of a SCARA robot and learning method achieves sufficiently precise motion to potentially enable automatic myocardial injection if similar results can be obtained in a real-world setting.
\end{abstract}


\end{frontmatter}
%===============================================================================

\section{Introduction}
\begin{figure*}[!t]
\centering
\includegraphics[width=0.8\linewidth]{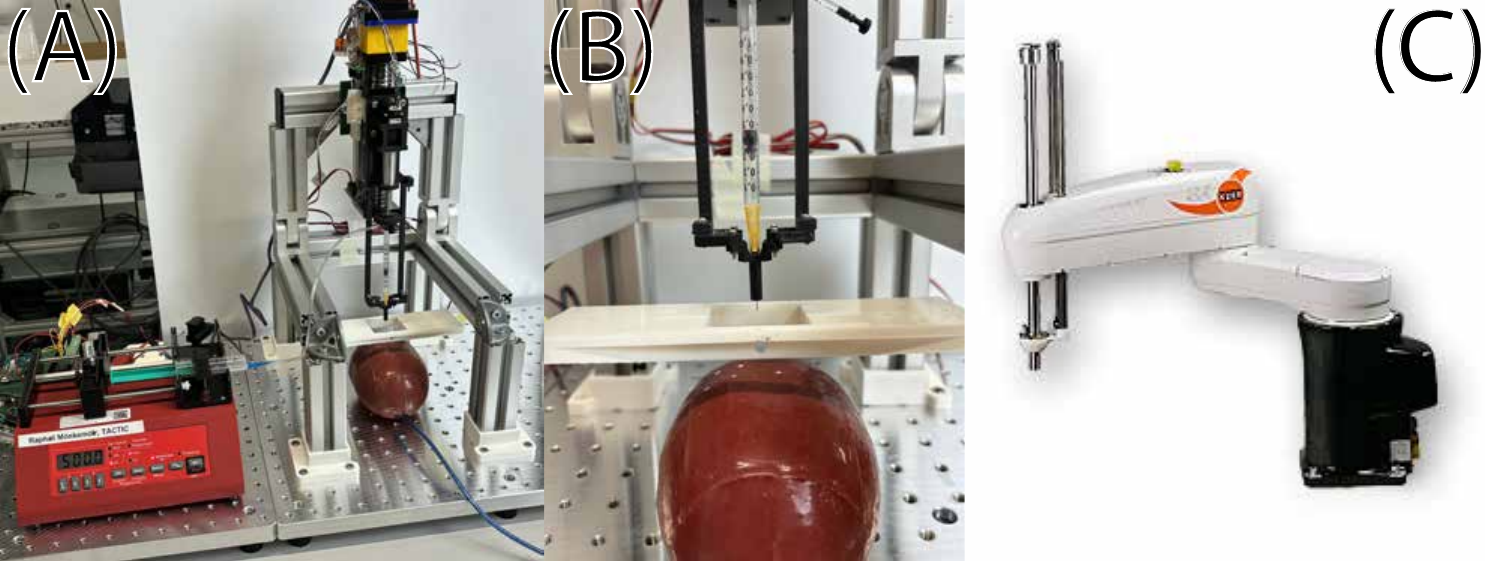}
\caption{This work considers the problem of automatically injecting stem cells into the myocardium using a robotic manipulator (A), which has to move a syringe with sub-millimeter precision. To meet this requirement, we propose Autonomous Iterative Motion Learning (AI-MOLE) for multi-input/multi-output systems and conduct a preliminary study using a SCARA robot (C), where the robot's endeffector is equipped with a syringe and the robot's two rotational degrees of freedom are utilized to position the syringe in the horizontal plane (B).}
\label{fig:heartinjection}
\end{figure*}

Stem cell therapy using human induced pluripotent stem cells (hiPCs) is a promising approach for treating the long-term effects of heart insufficiency. 
Here, direct injection into the myocardium [\cite{Mancuso.2020}] is advantageous because the stem cells are precisely delivered to the desired location.
In order to not require open sternatomy and, subsequently, decrease morbidity, direction injection should be carried out in a minimally invasive fashion.
However, the motion of the beating heart imposes great difficulties for the surgeon regarding the injection and dosing of cardiomyocytes.
Puncturing the wall or a vene would lead to a release of cells into the vascular system while injecting too quickly would lead to cell reflux and thus impaired cell retention [\cite{Teng.2006}].

To overcome these challenges, this work considers employing an applicator device that  automatically injects the cells into the heart wall.
Here, a robotic end effector is required that can move to multiple target locations with sub-millimeter precision.
At each target location, the robot has to move the syringe's needle tip to a desired depth to accurately inject the cells while a control algorithm compensates for the movement of the beating heart.

The precision and quality of the robotic motion is primarily determined by two components.
First, there is the choice of robot, and in this paper a  preliminary study is conducted to investigate whether the desired precision can be achieved using a SCARA robot, see Figure \ref{fig:heartinjection}.
Second, there is the choice of control approach.
Here, conventional, model-based control approaches could, in principle, be employed.
However, the robot's dynamics are not only affected by effects such as friction and backlash in the joints, but also by elasticity of the syringe and tubes, which are difficult to model.
Hence, a precise model of the robot's dynamics cannot be obtained, which inherently limits the precision of motion that can be achieved by model-based control approaches such as, e.g., \cite{Apgar2018, Golovin2019, Dong2005}.
As a consequence, learning control is employed in this work to overcome the lack of model information and achieve highly precise robotic motion [\cite{bristow2006survey, ahn2007iterative}].

One suited learning method is so called \emph{Autonomous Iterative Motion Learning} (AI-MOLE) [\cite{meindl2022bridging, meindl2024ai}] which solves reference tracking tasks by iteratively learning an input trajectory.
Unlike other learning approaches of Iterative Learning Control [\cite{seel2016iterative, freeman2012constrained}] or Reinforcement Learning [\cite{wu2023daydreamer,smith2022walk}] which either require a priori model information, manual parameter  tuning, or hours of system  interaction, AI-MOLE can be applied to unknown, nonlinear dynamics, only requires 5--15 trials of learning,  and  does not require manual parameter tuning.
Instead, the method self-reliantly determines necessary parameters and, hence, works in a plug-and-play fashion.
However, as presented in [\cite{meindl2024ai}], AI-MOLE is limited with respect to the SCARA robot considered in this work because the learning method has only been applied to single-input/single-output systems.
Hence, in this work we propose a methodological extension of AI-MOLE such that the method becomes applicable to multi-input/multi-output systems.

In summary, the contributions of this work are the following.
First, we propose an extension of AI-MOLE such that the method is applicable to systems with unknown, nonlinear, multi-input/multi-output dynamics.
Furthermore, the extended version of AI-MOLE is capable of self-reliantly determining necessary learning parameters, compensate coupling effects of multi-input/multi-output dynamics, and, hence, enable rapid learning in a plug-and-play fashion unlike existing learning approaches which either require a priori model information, manual parameter tuning, or excessive amounts of system interaction.
The proposed learning method is validated in a preliminary application study of a simplified SCARA robot.
Here, AI-MOLE learns to precisely track three different reference trajectories within only 15 trials per motion and without requiring any a priori model information or manual parameter tuning.
Furthermore, the simulation results indicate that the combination of the SCARA robot and AI-MOLE achieve sufficiently precise motion to potentially benefit the application of automated myocardial stem cell injections if similar results can be achieved in real world experiments.

\section{Problem Formulation}
The application problem motivating this work consists in  a SCARA robot moving a syringe with such precision that it becomes feasible to inject into a heart wall.
For this purpose, the robot's motor torques have to be carefully selected such that the robot's joint angles precisely track a desired reference trajectory despite the robot's dynamics being affected by unknown phenomena such as friction, backlash, or elasticity.
Hence, we consider the general problem of reference tracking via feedforward control in nonlinear, multi-input/multi-output systems with \emph{unknown} dynamics.
Note that the following formal description of the problem allows for reference tracking either in the robot's joint or task space.

Formally, consider a  multi-input/multi-output, discrete-time, repetitive system with a finite trial duration of $N\in\qPositiveNaturalNumbers$ samples, and, on trial $j\in\qPositiveNaturalNumbers$ and sample $n\in[1,N]$, input vector $\quvec_j(n)\in\qRealNumbers^R$, respectively output vector $\qyvec_j(n)\in\qRealNumbers^O$ with $R,~ O\in\qNaturalNumbers$.
The samples are collected in the so called input trajectory $\qu_j\in\qRealNumbers^{RN}$, respectively output trajectory $\qy_j\in\qRealNumbers^{ON}$, i.e., 
\begin{align}
    \qu_j &:= \begin{bmatrix}
    \quvec^\top_j(1) & \quvec^\top_j(2) & \dots & \quvec^\top_j(N)
    \end{bmatrix}^\top\,,
    \\
    \qy_j &:= \begin{bmatrix}
    \qyvec^\top_j(1) & \qyvec^\top_j(2) & \dots & \qyvec^\top_j(N)
    \end{bmatrix}^\top \,.
\end{align}
The system's state vector is denoted by $\qxvec_j(n)\in\qRealNumbers^M$, and the state dynamics are given by, $\forall j\in\qPositiveNaturalNumbers$,
\begin{align}\label{eq:statespace_dynamics}
    \qxvec_j(n+1) &:= \qf\left(\qxvec_j(n), \quvec_j(n)\right)\,,
    \\
    \label{eq:output_equation}
    \qyvec_j(n) &:= \qC\qxvec_j(n)\,,
\end{align}
where $\qf:\qRealNumbers^M\mapsto \qRealNumbers^M$ is the unknown, trial-invariant, nonlinear dynamics function and $\qC\in\qRealNumbers^{O\times M}$ is the known output matrix, i.e., the output vector is a linear combination of the state variables.

We assume that the desired motion is defined by a reference trajectory $\qr\in\qRealNumbers^{ON}$, and the control task consists in making the output trajectory $\qy_j$ precisely track the reference trajectory $\qr$. 
The learning task consists in updating the input trajectory $\qu_j$ from trial to trial such that the output trajectory $\qy_j$ converges to the desired reference trajectory $\qr$.
Tracking performance is measured by the error trajectory
\begin{equation}
    \forall j\in\qPositiveNaturalNumbers,\quad \qe_j:= \qr-\qy_j\,,
\end{equation}
and, ideally, the norm of $\qe_j$ should rapidly converge to zero and in a monotonic fashion.

\section{Proposed Method}
\begin{figure*}[!t]
\centering
\includegraphics[width=\linewidth, trim={0cm 0cm 0cm 6cm}, clip]{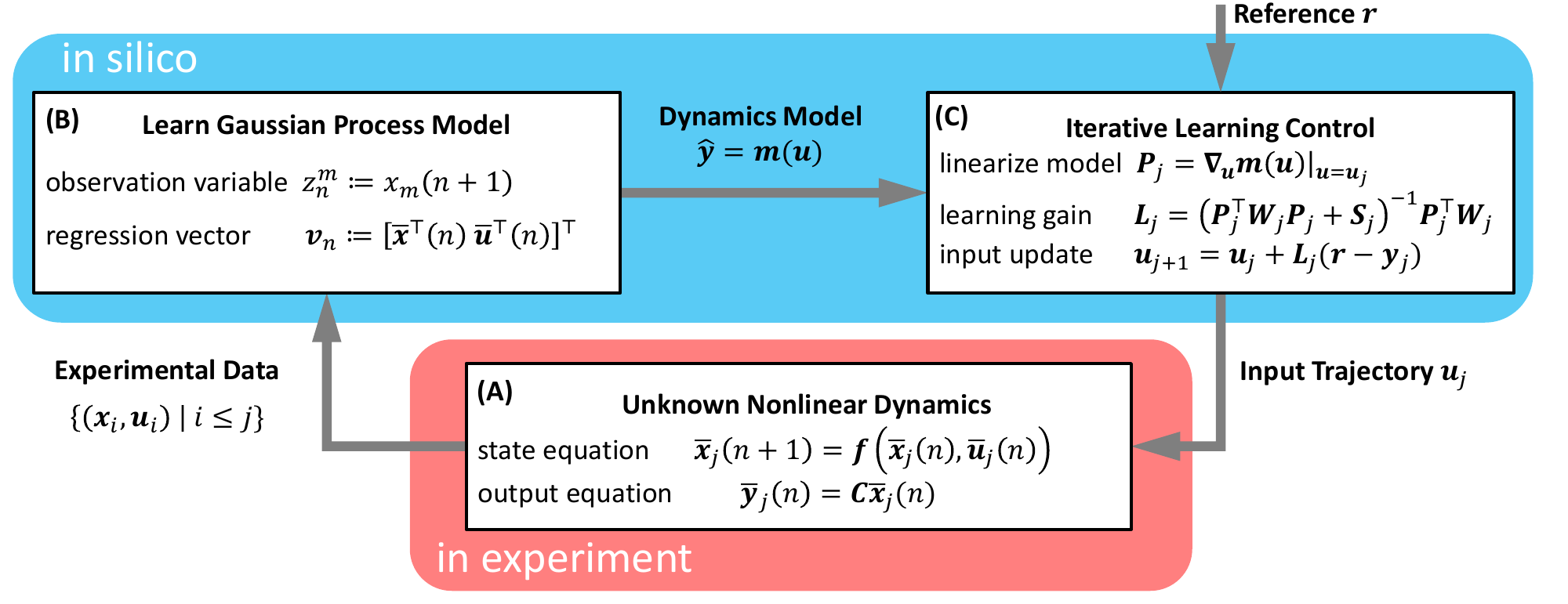}
\caption{Each of AI-MOLE's learning iterations consists of three steps. (A) The current input trajectory is applied to the unknown, multi-input/multi-output dynamics yielding a state trajectory. (B) The experimental data are then used to learn a GP model of the unknown dynamics. (C) Finally, AI-MOLE utilizes the GP model in an ILC update rule to compute the input trajectory of the next trial. The learning scheme is repeated until the output trajectory converges sufficiently close to the reference trajectory.}
\label{fig:method_overview}
\end{figure*}
We address the proposed problem by extending the method \emph{Autonomous Iterative Motion Learning} \linebreak(AI-MOLE), which was proposed in our previous work [\cite{meindl2024ai}], to be applicable to systems with multi-input/multi-output dynamics.
The proposed method initializes the input trajectory randomly and, from there onwards, each iteration consists of three steps, see Figure \ref{fig:method_overview}.
First, a GP model of the plant dynamics is identified using the experimental data of previous trials, see Section \ref{sec:model_learning}.
Second, the next-trial input trajectory is determined based on a norm-optimal Iterative Learning Control (ILC) update law using the GP model, see Section \ref{sec:input_learning}.
Third, the updated input trajectory is applied to the plant and the resulting data is in turn used to refine the GP model.
To enable plug-and-play application, the proposed method self-reliantly determines necessary parameters, see Section \ref{sec:autonomous_parameters}.
\subsection{Model Learning}\label{sec:model_learning}
To model the unknown dynamics, we utilize GPs and follow the fundamentals and notation as presented in [\cite{meindl2024ai}].
We propose a model, formally a function $\qm:\qRealNumbers^{RN}\mapsto \qRealNumbers^{ON}$, that predicts the plant's output trajectory $\qyest\in\qRealNumbers^{ON}$ based on an input trajectory $\qu\in\qRealNumbers^{RN}$, where the trial index is omitted to simplify notation.
We use GPs to model the state space dynamics \eqref{eq:statespace_dynamics}, whereby $M$ GPs are trained with each GP predicting the next sample of a respective state variable.
Hence,  the observation variable of the  $m^\mathrm{th}$ GP is defined as
\begin{equation}
    z^m_n := \left[\qxvec(n+1)\right]_m\,,
\end{equation}
and the regression vector consists of the current state sample $\qxvec$ and input sample $\quvec$, i.e.,
\begin{equation}\label{eq:regression_vector_is}
    \\qv_n := \begin{bmatrix}
    \qxvec(n)^\top & \quvec(n)
    \end{bmatrix}^\top\,.
\end{equation}
As kernel function, we employ a squared exponential kernel [\cite{RasmussenW06}] with $M$ length scales $\forall m\in[1,M], l_m\in\qRealNumbers$, i.e.,
\begin{multline}
    k(\qv, \hat{\qv})=\exp \left(-\frac{1}{2} \left(\qv-\hat{\qv}\right)^\top \boldsymbol{\Lambda}^{-2}\left(\qv-\hat{\qv}\right)\right) \quad \vert \\ \boldsymbol{\Lambda} = \mathrm{diag}\left(l\idx{1}, l\idx{2}, \dots, l_M\right)\,,
\end{multline}
whereby the length scales $l$ and measurement variance $\sigma^2_\omega$  are automatically determined via evidence maximization.
Furthermore, the training data are limited to the last $H\in\qNaturalNumbers$ trials to limit the model's computational requirements.

To predict the output trajectory $\qyest$ for a given input trajectory $\qu$, roll-out predictions are employed to predict the progression of the state vector $\qxvec$ over samples.
The progression of the output vector $\qyvec$ follows from $\qxvec$, the output matrix $\qC$, and the the output equation \eqref{eq:output_equation}.

Note that, if the system's state vector is unknown or not measured, the GP framework can also be utilized to model the input/output dynamics of the unknown dynamics which comes at the cost of a reduced speed of learning [\cite{meindl2024ai}].

\subsection{Input Learning}\label{sec:input_learning}
\begin{figure*}[!t]
\centering
\includegraphics[width=\linewidth]{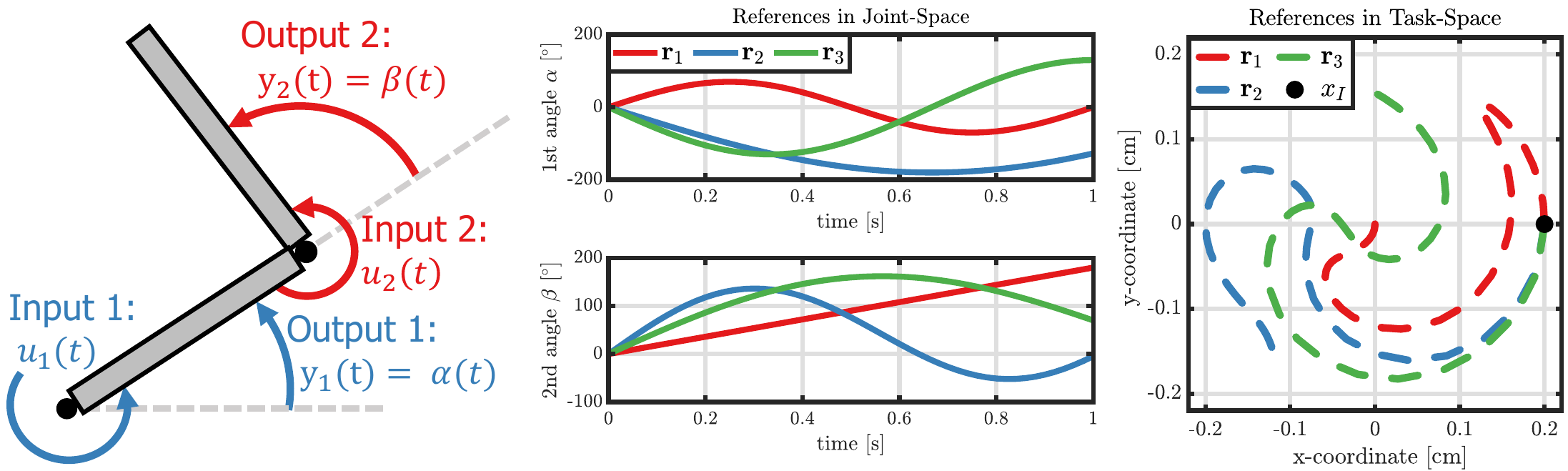}
\caption{AI-MOLE is evaluated using a simplified SCARA robot that consists of two links that rotate in the horizontal plane. AI-MOLE has to solve three different reference tracking tasks which respectively lead to dynamic motions that cover the robot's entire task space.}
\label{fig:simulation_task}
\end{figure*}
Once trained, the GP model is utilized in an ILC update law to compute the input trajectory of the next trial.
Here, AI-MOLE employs the linear, trial-varying update law of the form 
\begin{equation} \label{eq:input_update_law}
    \quad \qu_{j+1} = \qu_j + \qdeltau_j \quad \vert ~ \qdeltau_j=\qL_j\qe_j\,,
\end{equation}
where $\qL_j\in\qRealNumbers^{ON\times RN}$ is the trial-varying learning gain matrix.
To select the learning gain matrix $\qL_j$, the nonlinear GP model $\qm$ is linearized at the current input trajectory $\qu_j$, i.e,
\begin{equation}
   \qP_j := \qgrad_\qu \qm(\qu)\vert_{\qu=\qu_j}\,,
\end{equation}
and linearity of the dynamics is assumed, i.e.,
\begin{equation}
    \hat{\qy}_{j+1} = \qy_j + \qP_j \qdeltau_j\,.
\end{equation}
Given the linearization $\qP_j$, the learning gain matrix is designed via norm-optimal ILC [\cite{bristow2006survey}], i.e.,
\begin{equation}\label{eq:learning_gain_design}
    \qL_j = \left(\qP_j^\top \qW_j\qP_j + \qS_j\right)^{-1}\qW_j \qP_j^\top\,,
\end{equation} 
where $\qW_j\in\qRealNumbers^{ON\times ON}, \qS_j\in\qRealNumbers^{RN\times RN}$ are symmetric, positive-definite weighting matrices.
Note that AI-MOLE can also employ other model-based ILC approaches such as, e.g., gradient-based ILC [\cite{huo2020data}].

\subsection{Autonomous Parameterization}\label{sec:autonomous_parameters}
To realize plug-and-play application, AI-MOLE self-reliantly determines the necessary parameters by the following procedure.

First, the initial input trajectory $\qu\idx{1}$ is considered which is applied on the very first trial.
Note that $\qu\idx{1}$ generally is a robust parameter, i.e., a variety of choices of $\qu\idx{1}$ lead to successful learning [\cite{meindl2022bridging}].
The input trajectory $\qu\idx{1}$ is selected by first determining the reference trajectory's largest significant frequency $f\idx{0}$ based on the reference's frequency spectrum and  subsequently designing a zero-phase low-pass filter $\qf\idx{LP}$ with the cut-off frequency $f\idx{0}$.
Next, AI-MOLE applies the low-pass filter $\qf\idx{LP}$ to a zero-mean normal distribution with covariance $\sigma^2\idx{I}\qI$, and the initial input trajectory is drawn from the resulting distribution, i.e.,
\begin{equation}\label{eq:initial_input}
    \qu\idx{1} \sim \qf\idx{LP}\left(\mathcal{N}(\qvec{0}, \sigma^2\idx{I}\qI)\right)\,.
\end{equation}
To provide sufficient excitation, the input variance $\sigma^2\idx{I}$ is set to a small value such that the system's output exceeds the level of the measurement noise.
If necessary, the input variance $\sigma^2\idx{I}$ is iteratively determined by trivially incrementing the input variance until the system is sufficiently excited.

The ILC input update requires the weights $\qW_j$ and $\qS_j$ which are autonomously determined based on the linearized model $\qP_j$.
Here, multi-input/multi-output systems pose a particular challenge because the weights have to accommodate possible cross-coupling effects between the various input and output variables.
For this purpose, let $\qP^{ro}_j\in\qRealNumbers^{N\times N}, ~o\in[1,O], ~r\in[1,R]$, denote the matrix that consists of the elements of $\qP_j$ such that $\qP^{ro}_j$ maps the effect of the $r^\mathrm{th}$ input variable's trajectory on the $o^\mathrm{th}$ output variable's trajectory.
Now, the weighting matrices are given by
\begin{equation}\label{eq:weights}  
\qW_j = \qI \otimes \bar{\qW}_j, \quad \qS_j =\qI\otimes  \bar{\qS}_j\,,
\end{equation}
where $\otimes$ is the Kroenecker product and
\begin{multline}
\bar{\qW}_j = \mathrm{diag}\left(
\qNorm{\begin{bmatrix}
\qP^{11}_j & \dots & \qP^{1r}_j
\end{bmatrix}}^{-1}
~~ \dots ~~ \right. \\
\left.\qNorm{\begin{bmatrix}
\qP^{m1}_j & \dots & \qP^{mr}_j
\end{bmatrix}}^{-1}
\right)\,,
\end{multline}
\begin{multline}
\bar{\qS}_j = \mathrm{diag}\left(
\qNorm{\begin{bmatrix}
\qP^{11}_j & \dots & \qP^{m1}_j
\end{bmatrix}}
~~ \dots ~~ \right. \\
\left.\qNorm{\begin{bmatrix}
\qP^{1r}_j & \dots & \qP^{mr}_j
\end{bmatrix}}
\right)\,.
\end{multline}
To reduce AI-MOLE's computational requirements, the number of trials that are used to train the GP model is set to $H=3$ because the complexity of GPs is cubic with respect to the number of data points.

\section{Simulation Results}
\begin{figure*}[!t]
\centering
\includegraphics[width=\linewidth]{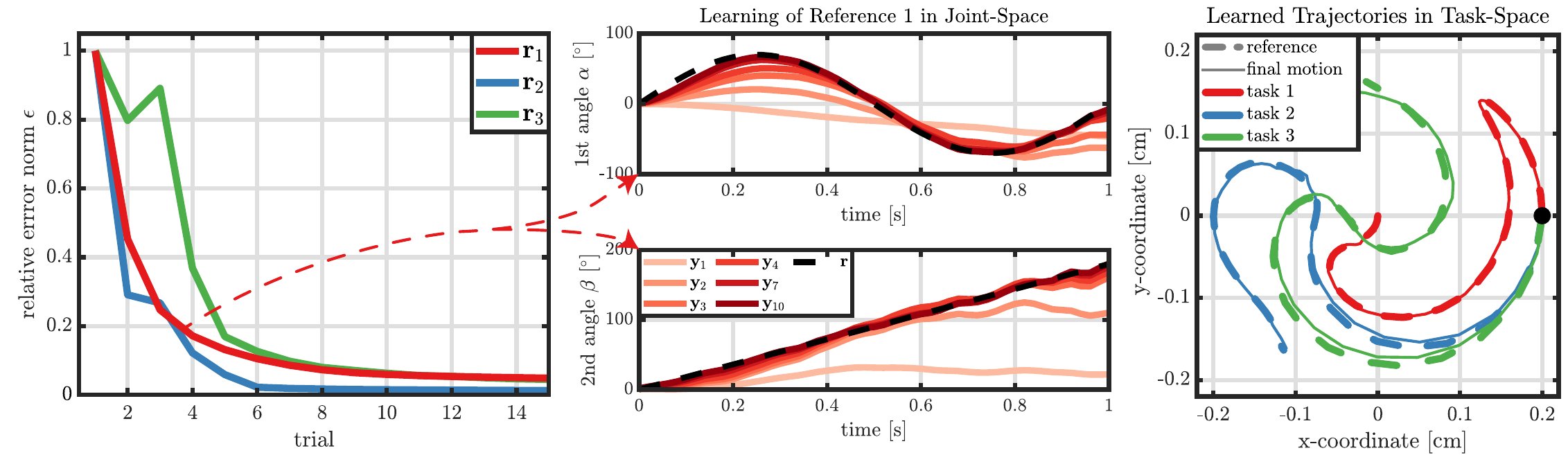}
\caption{The simulation results of AI-MOLE learning to track three different reference trajectories. Despite the unknown, multi-input/multi-output dynamics, AI-MOLE achieves satisfying tracking performance within 10--15 trials without any manual parameter tuning, and the final motions are sufficiently precise to possibly enable automated myocardial injection.}
\label{fig:simulation_results}
\end{figure*}

In this section, AI-MOLE's capability of learning to have a SCARA robot precisely perform desired motions is going to be investigated.
For this purpose, we employ a simplified dynamics model of a SCARA robot, in which only the first two rotational degrees of freedom are considered because the latter pose the most challenging, nonlinear component of a SCARA robot's dynamics.
The simplified robot consists of two links that rotate in the horizontal plane, see Figure \ref{fig:simulation_task}.
The angle of the first link is denoted by $\alpha\in\qRealNumbers$ and the angle of the second link relative to the first link is denoted by $\beta\in\qRealNumbers$.
Both of the robot's joints are driven by motors, the first link's motor torque is denoted by $u\idx{1}$, and the second link's motor torque is denoted by $u\idx{2}$.
The input, state, and output vectors are hence given by
\begin{equation}
\qyvec(n) = \begin{bmatrix}
\alpha(n) & \beta(n)
\end{bmatrix}^\top\,,
\end{equation}
\begin{equation}
\quvec(n) = \begin{bmatrix}
u(n) & u(n)
\end{bmatrix}^\top\,,
\end{equation}
\begin{equation}
\qxvec(n) = \begin{bmatrix}
\alpha(n) & \beta(n) & \dot{\alpha}(n) & \dot{\beta}(n)
\end{bmatrix}^\top\,,
\end{equation}
and the output matrix follows with
\begin{equation}
\qC = \begin{bmatrix}
\qI& \qmat{0}
\end{bmatrix}\,.
\end{equation}
To simulate the robot, the  nonlinear dynamics were derived using the Lagrange formalism, implemented in a black-box simulation using MATLAB, and necessary mechanical parameters were identified via the CAD model.
However, the learning method is provided \emph{no} model information, and, for the remainder of these simulations, the robot's dynamics are assumed to be \emph{unknown}.

To validate AI-MOLE, the robot has to learn  three different desired motions, which are encoded as reference trajectories in joint space, and learning performance is judged based on the normalized error norm
\begin{equation}
\epsilon_j:= \frac{\qNorm{\qr -\qy_j}}{\qNorm{\qr-\qy\idx{1}}}\,.
\end{equation}
The reference trajectories greatly differ in amplitudes and frequencies and lead to highly dynamic motions that cover the robot's entire task space, see Figure \ref{fig:simulation_task}.

To now learn the three motions, we employ AI-MOLE.
Here, for each of the three references, an initial input trajectory is randomly generated according to \eqref{eq:initial_input} and applied to the robot.
AI-MOLE employs the input and resulting state trajectory to learn the first GP model of the unknown dynamics and uses the model to update the input trajectory according to \eqref{eq:input_update_law} with the \emph{autonomously} determined parameters \eqref{eq:weights}.
Afterwards, the learning scheme is repeated for a total of 15 iterations.

The results depicted in Figure \ref{fig:simulation_results} show that AI-MOLE is capable of rapidly learning all three of the desired motions whereby the error norm drops below 20\% of the initial value within five trials and converges to a value close to zero within 10--15 trials.
Furthermore, the error norms are decreasing monotonically for all references and trials, except on a single trial.
The progression of the output trajectories in task and joint space show that the finally learned motions precisely track the desired motions and learning, hence, succeeded within just 15 trials.
The results further demonstrate that AI-MOLE is capable of learning to solve reference tracking tasks for multi-input/multi-output systems in a truly autonomous fashion because AI-MOLE did not require any \emph{manual} parameter tuning.

\section{Conclusion}
In this work, the application of automated myocardial injection has been considered, and the combination of a SCARA robot and a learning control approach has been investigated.
For this purpose, an extension of the method Autonomous Iterative Motion Learning (AI-MOLE) has been proposed to enable application to multi-input/multi-output systems.
The proposed method solves reference tracking tasks in multi-input/multi-output systems with unknown, nonlinear dynamics by iteratively updating an input trajectory.
In each iteration, AI-MOLE uses experimental data to train a GP model of the unknown, nonlinear, multi-input/multi-output dynamics.
Then, AI-MOLE uses the model in an ILC update law to update the input trajectory, whereby necessary learning parameters are determined automatically.

The key advantages of the proposed learning methods are that the method works in a plug-and-play fashion and requires neither manual parameter tuning nor a priori model information.
Furthermore, the method is capable of learning to solve reference tracking tasks from scratch in as little as 10--15 trials.

The proposed method was validated by a simulation study of a simplified SCARA robot which has to perform three dynamic motions.
The simulation results confirm that AI-MOLE is capable of achieving highly precise motion control within a small number of trials and without any a priori model information or manual parameter tuning.
Furthermore, the results indicate that the learning method achieves sufficiently precise motion to possibly enable automated injection into the myocardium if similar performance can be achieved in a real-world environment.
Previous work has shown that AI-MOLE achieves similar performance in simulated and real-world environments \cite{meindl2022bridging}.

While the presented results are promising, they are currently subject to the following limitations.
The proposed method was only validated in simulation and the simulated MIMO system only had two input and output variables. 
Hence, future work is going to to validate AI-MOLE on a real-world MIMO system and investigate the method's performance on system's with a larger number of input and output variables.
Also, the proposed method is going to be compared with other learning approaches.

\bibliography{ifacconf}

\end{document}